\documentclass[10pt,twocolumn,letterpaper]{article}
\pdfoutput=1
\usepackage[pagenumbers]{cvpr} 

\usepackage{graphicx}
\usepackage{amsmath}
\usepackage{amssymb}
\usepackage{booktabs}

\usepackage[pagebackref,breaklinks,colorlinks]{hyperref}

\usepackage[capitalize]{cleveref}
\crefname{section}{Sec.}{Secs.}
\Crefname{section}{Section}{Sections}
\Crefname{table}{Table}{Tables}
\crefname{table}{Tab.}{Tabs.}

\def\cvprPaperID{34} 
\def\confName{CVPR}
\def\confYear{2024}

\begin{document}

\title{InstructRL4Pix: Training Diffusion for Image Editing by Reinforcement Learning}

\author{Tiancheng Li\thanks{These authors contributed equally to this work.}\\
South China University of Technology\\
{\tt\small ltc20020913@gmail.com}
\and
Jinxiu Liu\footnotemark[1]\\
South China University of Technology\\
{\tt\small jinxiuliu0628@gmail.com}
\and
Huajun Chen\\
South China University of Technology\\
{\tt\small ftchenhuajun@mail.scut.edu.cn}
\and
Qi Liu\thanks{Corresponding author.}\\
South China University of Technology\\
{\tt\small drliuqi@scut.edu.cn}
}


\maketitle


\begin{figure*}
  \includegraphics[width=\textwidth]{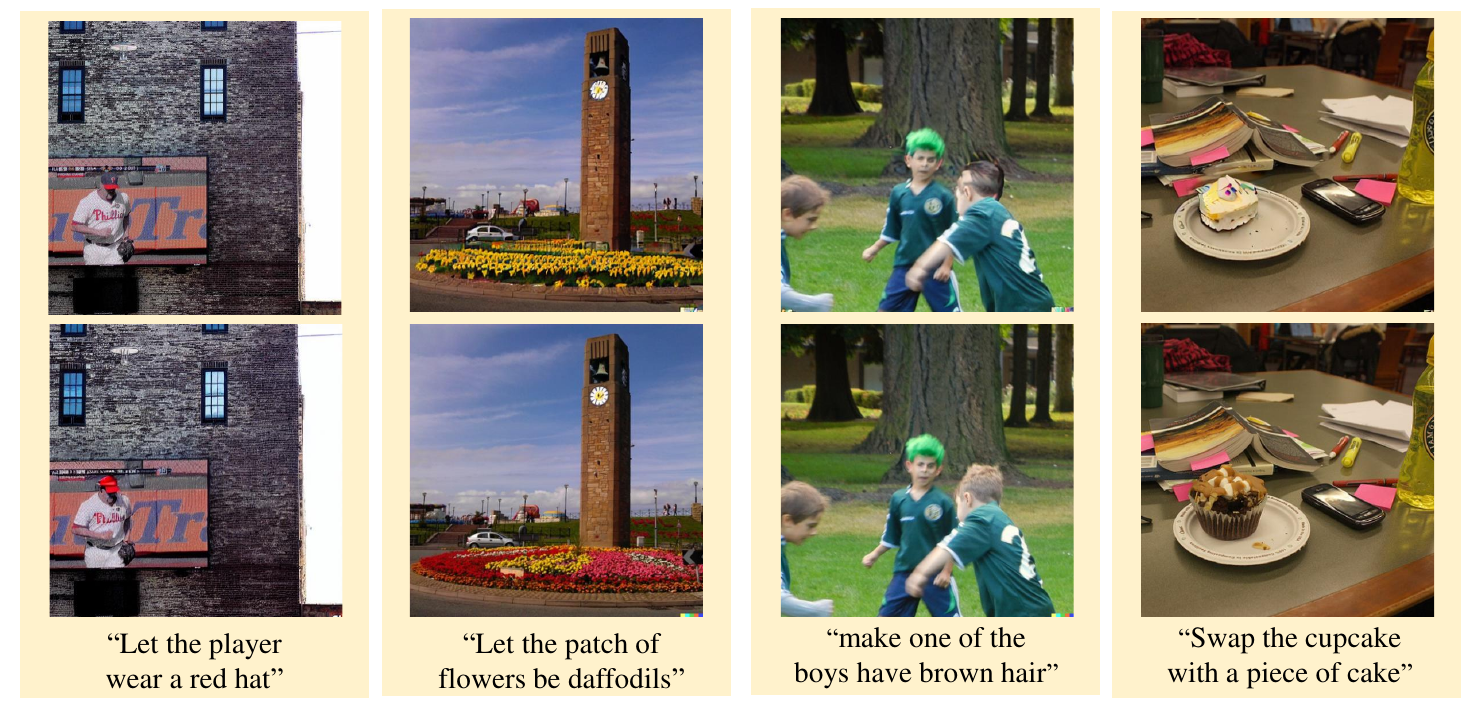}
  \caption{We introduce Reinforcement Learning Guided Image Editing Method(InstructRL4Pix) to unsupervised optimize instruction-based image editing for various editing tasks. The bottom is the edit instruction, the middle is the input image, and the top is the output image after InstructRL4Pix editing.}
  \label{fig:example}
\end{figure*}

\begin{abstract}
   Instruction-based image editing has made a great process in using natural human language to manipulate the visual content of images. However, existing models are limited by the quality of the dataset and cannot accurately localize editing regions in images with complex object relationships. In this paper, we propose Reinforcement Learning Guided Image Editing Method(InstructRL4Pix) to train a diffusion model to generate images that are guided by the attention maps of the target object. Our method maximizes the output of the reward model by calculating the distance between attention maps as a reward function and fine-tuning the diffusion model using proximal policy optimization (PPO). We evaluate our model in object insertion, removal, replacement, and transformation. Experimental results show that InstructRL4Pix breaks through the limitations of traditional datasets and uses unsupervised learning to optimize editing goals and achieve accurate image editing based on natural human commands.
\end{abstract}

\section{Introduction}

Instruction-based image editing has made a great process in using natural human language to manipulate the visual content of images. Diffusion models\cite{Ho_Jain_Abbeel_2020} have been widely used in image editing\cite{Kim_Kwon_Ye_2022}.Instruction-based editing\cite{El-Nouby_Sharma_Schulz_Hjelm_Asri_Kahou_Bengio_Taylor_2018}\cite{Li_Qi_Lukasiewicz_Torr_2020}\cite{Fu_Wang_Grafton_Eckstein_Wang_2020} enables humans to give editing commands directly to images, rather than relying on detailed descriptions\cite{hertz2022prompttoprompt}\cite{Mokady_2023_CVPR}\cite{Kawar_2023_CVPR}. In order to obtain the training data for edit task, InstructPix2Pix\cite{Brooks_2023_CVPR} use GPT-3\cite{Brown_Mann_Ryder_Subbiah_Kaplan_Dhariwal_Neelakantan_Shyam_Sastry_Amanda_et_al._2020} and Prompt-to-Prompt\cite{hertz2022prompttoprompt} to generate datasets and train the diffusion model on these dataset. However, the capabilities of GPT-3 and Prompt-to-Prompt limit the quality of this dataset, making existing image editing methods less effective in editing images with complex relationships.

Proximal Policy Optimization (PPO)\cite{Schulman_Wolski_Dhariwal_Radford_Klimov_2017} is a reinforcement learning algorithm that directly optimizes the policy to find the policy that maximizes the expected payoff.DDPO\cite{black2024training} models the denoising process of a diffusion model as a multi-step Markov Decision Process (MDP) and optimizes the diffusion model using the PPO.

Inspired by DDPO, we propose the Reinforcement Learning Guided Image Editing Method(InstructRL4Pix) to train a diffusion model to generate images that are guided by the attention maps of the target object. Our method maximizes the output of the reward model by calculating the distance between attention maps as a reward function and fine-tuning the diffusion model using proximal policy optimization (PPO). To train the InstructRL4Pix, we adopt MagicBrush\cite{NEURIPS2023_64008fa3} as our pre-training dataset. We evaluate our model in object insertion, removal, replacement, and transformation. 
Our model realizes accurate local editing based on human commands while preserving the features of the original image. Our contributions are summarised as follows:

\begin{itemize}
    \item We propose the Reinforcement Learning Guided Image Editing Method (InstructRL4Pix), which can successfully handle complex editing scenarios and no longer limited by the GPT-3 and Prompt-to-Prompt features.
    \item We introduce a training method that allows for training without touching the edited image and modeling the similarity between the attention maps to guide the editing process.
\end{itemize}


\section{Related Works}
The beginning text-guided image editing uses GAN to improve the flexibility and accessibility of image visual content based on natural instructions.\cite{10.1145/3422622}\cite{Reed_Akata_Yan_Logeswaran_Schiele_Lee_2016}. The diffusion model\cite{Ho_Jain_Abbeel_2020} has greater controllability, which includes training methods\cite{Zhang_2024_CVPR}\cite{yildirim2023instinpaint},  testing-time finetuning methods\cite{Kawar_2023_CVPR}\cite{Mokady_2023_CVPR}, and training \& finetuning free methods\cite{10.1145/3588432.3591513}\cite{Avrahami_Lischinski_Fried_2022}. Unlike them, instruction-based image editing accepts human commands directly, including object insertion, removal, replacement, and transformation. Recent approaches have improved the editing quality in terms of synthesizing input target instruction triples \cite{Brooks_2023_CVPR}. However, their editing ability is either limited by the GPT-3\cite{Brown_Mann_Ryder_Subbiah_Kaplan_Dhariwal_Neelakantan_Shyam_Sastry_Amanda_et_al._2020} and Prompt-to-Prompt\cite{hertz2022prompttoprompt} capabilities. In this paper, we calculate the distance between attention maps as a reward function and fine-tuning the diffusion model using proximal policy optimization (PPO), which accurately limits the editing area of the image and improves the model's editing performance on complex images.

\section{Method}
We propose the Reinforcement Learning Guided Image Editing Method(InstructRL4Pix) to train a diffusion model to generate images that are guided by the attention maps of the target object. We calculate the distance between attention maps as a reward function and fine-tuning the diffusion model using proximal policy optimization (PPO).

\subsection{Reward Function\label{sec3.1}}
\paragraph{Attention Map Loss.}
Given an input image $ \mathcal{V} $, mask image $ \mathcal{M} $ and an instruction $ \mathcal{X} $, we want the editing model to generate an output image $ \mathcal{O} $ that contains the target object specified by $ \mathcal{X} $. We use the mask image $ \mathcal{M} $ to generate the ground-truth attention map $a_{1}$ that indicates the real location of the target object in $ \mathcal{V} $. 

By computing the cross-attention between the encoded text and the intermediate features of the denoisers $\epsilon_{\theta}$, the pre-trained InsPix2Pix editing model generates the reference attention map $a_{1}$ from $ \mathcal{X} $. 
\begin{align}
\text{Attention}(Q, K, V) &= M \cdot V \\
\text{where } M &= \text{softmax}\left(\frac{QK^T}{\sqrt{d}}\right) 
\end{align}
Here, we are primarily concerned with the cross-attention map $M$, which is observed that it is closely related to the structure of the image\cite{hertz2022prompttoprompt}.$M_{ij}$ represents the weight of the value of the $j-th$ token on the pixel $i$. In addition, the cross-attention mask is specific to a certain time step, and we get a different attention mask $M_t$ at each time step $t$. We define a1 as the average of all tokens over the entire sampling time.
\begin{equation}
a_1 = \frac{1}{N} \sum_{n=1}^{N} (\frac{1}{T} \sum_{t=0}^{T-1} M_t)
\end{equation}
Where N is the number of tokens in the instruction $ \mathcal{X} $, T is the diffusion steps. 

The attention map loss is defined as the cosine similarity between a1 and a2, which measures how well the editing model can align the target object with the instruction.
\begin{equation}
L_{att} = \frac{a_1 \cdot a_2}{\|a_1\| \|a_2\|}
\end{equation}
The attention map loss can be seen as a proxy for the perceptual quality of the output image, as it encourages the editing model to preserve the semantic content of the input image and modify only the relevant region.

\begin{figure}
  \includegraphics[width=0.5\textwidth]{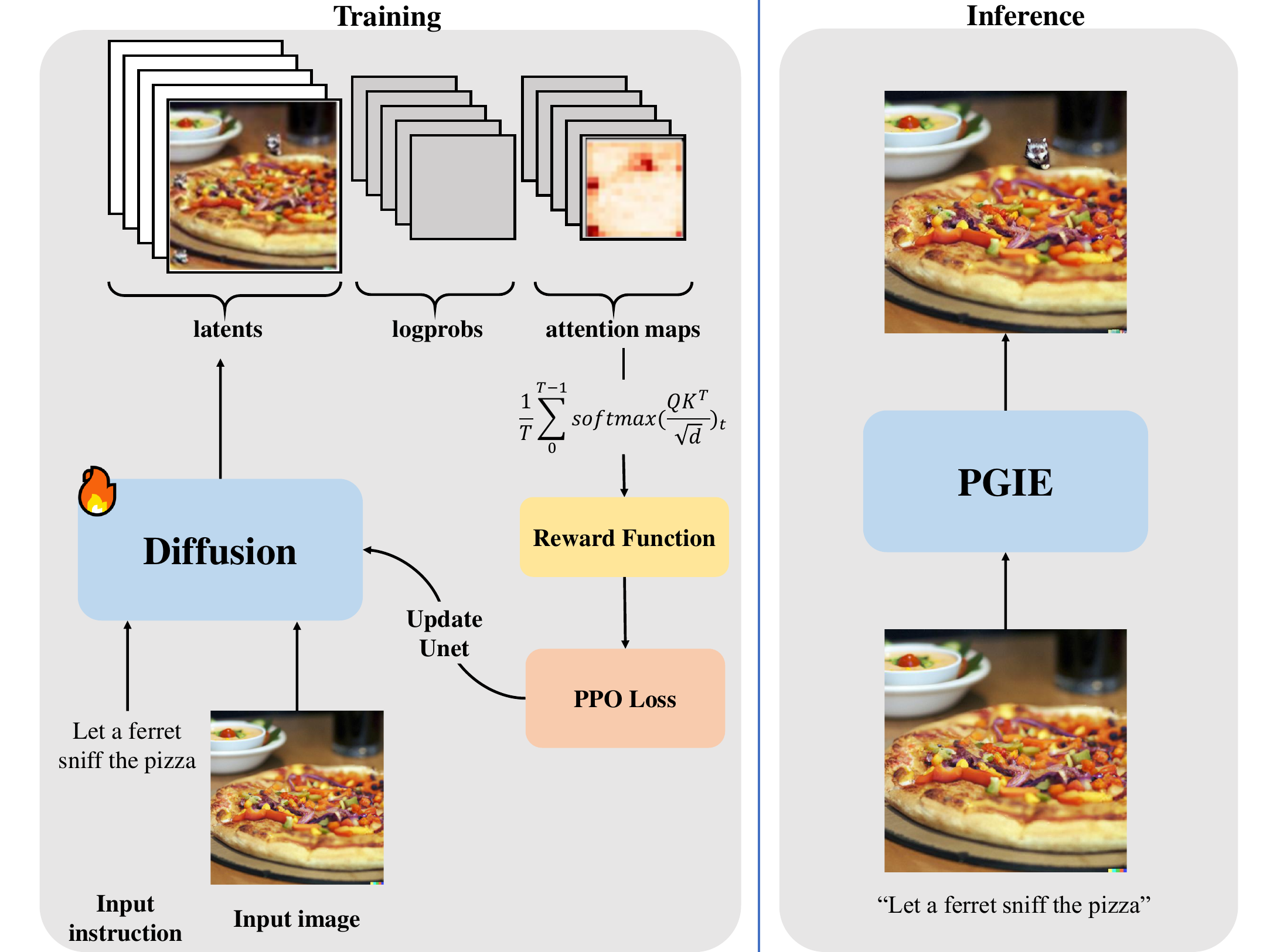}
  \caption{Overview of Reinforcement Learning Guided
Image Editing Method(InstructRL4Pix) to train a diffusion model to generate images that are guided by the attention maps of the target object. InstructRL4Pix breaks through the limitations of traditional datasets and uses unsupervised learning to optimize editing goals and achieve accurate image editing based on natural human commands.}
  \label{fig:structure}
\end{figure}

\paragraph{Clip Loss.}
Clip Loss is a regularization term that penalizes the editing model for generating output images that are too different from the input images in terms of pixel values. The Clip Loss is defined as the mean absolute error (MAE) between the input image and the output image, clipped by a threshold $\tau$:

\begin{equation}
L_{clip} = \text{MAE}(\mathcal{V}, \mathcal{O}) \cdot \mathbb{I}(\text{MAE}(\mathcal{V}, \mathcal{O}) > \tau)
\end{equation}
where $\mathbb{I}$ is an indicator function. The Clip Loss prevents the editing model from making drastic changes to the input image.

\paragraph{Total Reward Function} The total reward function is a weighted sum of these losses:
\begin{equation}
r(x_0, \mathcal{V}, \mathcal{X}) = L_{att} + \alpha \cdot L_{clip}
\end{equation}
where $\alpha$ is the hyperparameter that controls the trade-off between different objectives. We optimize this loss using PPO algorithm as described in Section \ref{sec3.2}.

\subsection{Reinforcement Learning Guided Image Editing\label{sec3.2}}

\paragraph{Modeling editing model as Multi-Step MDPs.}
The Markov decision process(MDP) can be formalized as ($ \mathcal{S} $,$ \mathcal{A} $,$\rho_0$,$ \mathcal{P} $,$ \mathcal{R} $), where $ \mathcal{S} $ is the state space, $ \mathcal{A} $ is the action space, $\rho_0$ is the initial state distribution, $ \mathcal{P} $ is the state transfer matrix, and $ \mathcal{R} $ is the reward function. The optimization objective of reinforcement learning is to maximize the cumulative payoff of the strategy:
\begin{equation}
\mathcal{J}_{RL} (\pi)= \mathbb{E}_{\tau \sim p(\tau|\pi)} \sum_{t=0}^{T} \mathcal{R}(s_t,a_t)
\end{equation}

We formulate the image editing task as an MDP, where the state is the input image$\mathcal{V}$, the action is a noise mask applied to the image, and the reward is the function mentioned in Section \ref{sec3.1}. 

Following DDPO, we align the denoising process of the diffusion model based editing model with that of the MDP so that the optimization objective $\mathcal{J}_{RL} (\pi)$ is Equivalent to $\mathcal{J} (\pi)$:
\begin{equation}
\mathcal{J} (\pi)= \mathbb{E}_{\mathcal{V} \sim p(\mathcal{V}), \mathcal{X} \sim p(\mathcal{X}), x_0 \sim p_\theta(x_0|(\mathcal{V},\mathcal{X})} [r(x_0, \mathcal{V}, \mathcal{X})] 
\end{equation}

\paragraph{Policy Gradient Estimation}
To optimize , it is then necessary to estimate its gradient. We use PPO for gradient estimation of $\mathcal{J} (\pi)$:
$$
\nabla \mathcal{J} (\pi) = \mathbb{E} \bigg[ \& \bigg( \text{clip} \bigg( \frac{p_\theta(x_{t-1}|x_t,\mathcal{V}, \mathcal{X})}{p_{\theta_{\text{old}}}(x_{t-1}|x_t,\mathcal{V}, \mathcal{X})}, 1-\epsilon, 1+\epsilon \bigg) \\
$$
\begin{equation}  
\& \nabla_\theta \log p_\theta(x_{t-1}|x_t,\mathcal{V}, \mathcal{X}) \cdot r(x_0, \mathcal{V}, \mathcal{X}) \bigg) \bigg]
\end{equation}
Once the estimates are obtained, an optimization algorithm can be applied to optimize the parameters of the diffusion model to maximize.

\section{EXPERIMENT}
\subsection{ Implementation Details}
\paragraph{Datasets and Evaluation Metrics.}
We use MagicBrush \cite{NEURIPS2023_64008fa3} as our pre-training data. It comprises 10K (source image, instruction, target image) triples. However, we only use the source image, mask image and instruction from the dataset, without using the target image, which avoids the impact on the model training due to the quality of the target image. 
For a comprehensive evaluation, we consider various editing aspects. We selected 2 existing instruction-based image editing models: InstructPix2Pix\cite{Brooks_2023_CVPR} and Fine-tuned InsPix2Pix on MagicBrush\cite{NEURIPS2023_64008fa3} to compare with our results. All methods are evaluated on
MagicBrush test set. We use the L1 and L2 to measure the standard pixel difference between the generated image and ground truth image. Additionaly, CLIP-T \cite{Ruiz_2023_CVPR} is used to measure text-image alignment by the cosine similarity between the local description and the CLIP embedding of the generated image. Finally, SSIM\cite{Wang_Bovik_Sheikh_Simoncelli_2004} and PSNR\cite{Korhonen_You_2012} are used to measure the quality of the generated image compared to the original image.
\paragraph{Baselines} We treat InsPix2Pix \cite{Brooks_2023_CVPR},  a diffusion model for instruction-based image editing, as our baseline. 
\paragraph{Training Details} We treat DDPO\cite{black2024training} as our basecode. The instruction-based diffusion model $\mathcal{F}$ is initialized from finetuned-insPix2Pix of MagicBrush\cite{NEURIPS2023_64008fa3}. We train the UNet with low-rank adaptation (LoRA)\cite{J._Shen_Wallis_Allen-Zhu_Li_Wang_Chen_2021}. We adopt AdamW\cite{Loshchilov_Hutter_2017} with a batch size of 8  to optimize InstructRL4Pix. The learning rates of the $\mathcal{F}$ is 2e-4 and the hyperparameter $\alpha$ is -1, respectively. All experiments are conducted in PyTorch on 4 3090 GPUs.
\begin{figure}
  \includegraphics[width=0.5\textwidth]{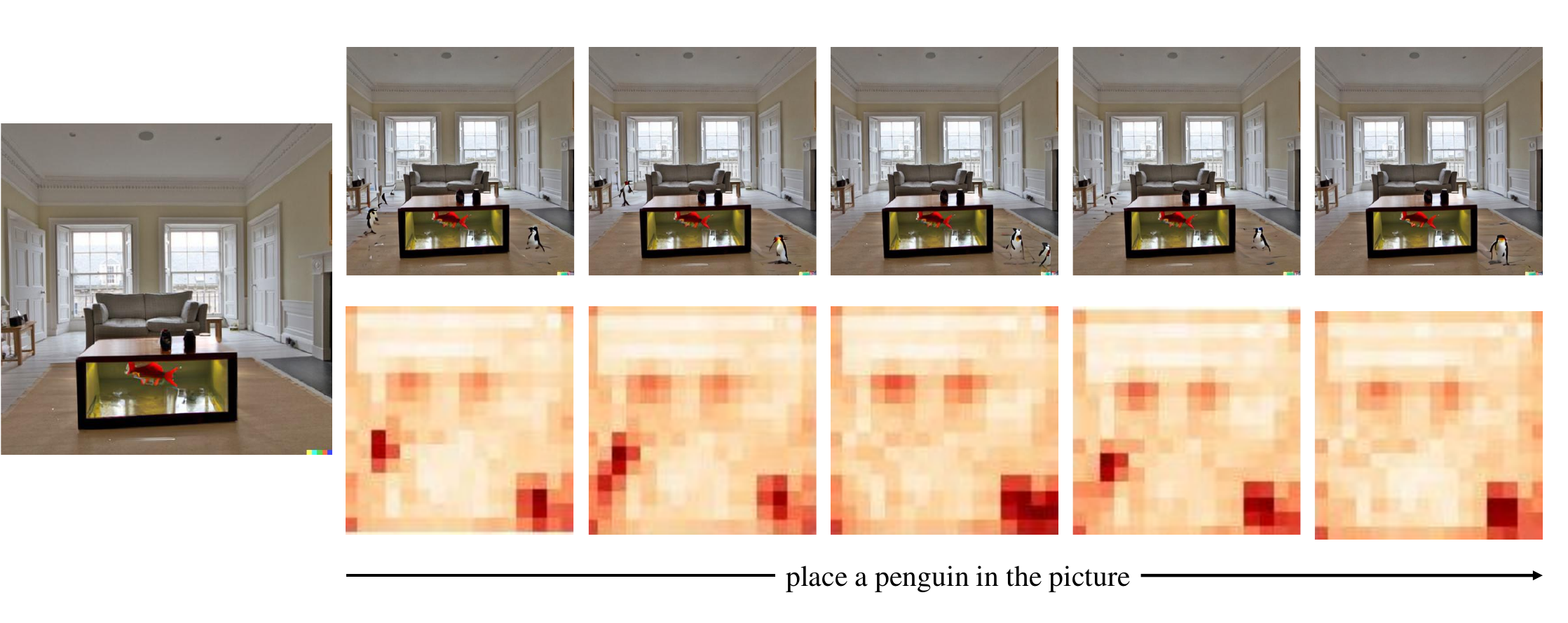}
  \caption{Sample progressions of the same cue and random seeds during training. The attention map of the samples will tend to localize more faithfully to the correct editing region.}
  \label{fig:process}
\end{figure}
\subsection{Evaluations}
Then we evaluate the ability of InstructRL4Pix to fine-tune Unet. The progression of finetuning is depicted in Figure \ref{fig:process}. Qualitatively, during training, the attention map of the samples will tend to localize more faithfully to the correct editing region. For example, we use edit instruction "place a penguin in the picture", during training, the attention map of the samples goes from being initially localized to the left and right to being specified to be located on the right, and the editing effect goes from increasing the penguins on both the left and right to increasing them only on the right. After fine-tuning, our model has the ability to accurately edit complex datasets.
 \begin{table}[h]
  \caption{Quantitative study of image editing baselines on MagicBrush test set. The best results are marked in bold.}
  \label{tab:magicbrush}
  \small 
  \setlength{\tabcolsep}{3pt} 
  \begin{tabular}{@{}cccccl@{}}
    \toprule
    Methods &L1$\downarrow$ &L2$\downarrow$&CLIP-T$\uparrow$&SSIM$\uparrow$&PSNR$\uparrow$\\
    \midrule
    InstructPix2Pix & 0.1122& 0.0371&\textbf{0.2764}&0.6840&19.9091\\
    MagicBrush & 0.0964& 0.0353&0.2754&0.7046&23.6468\\
    InstructRL4Pix & \textbf{0.0608}& \textbf{0.0189}&0.2391&\textbf{0.7978}&\textbf{24.2575}\\
  \bottomrule
\end{tabular}
\end{table}
\subsection{Comparison}
We first evaluate all methods on existing instruction-based image editing tasks: InstructPix2Pix, fine-tuned InsPix2Pix on MagicBrush and our InstructRL4Pix. all methods are evaluated on MagicBrush test set.
Table \ref{tab:magicbrush} shows the results of the instruction-based image editing methods. The results show that InstructRL4Pix has a lower L1 and L2 and higher SSIM and PSNR compared with the existing methods, which is due to the fact that we use Clip-loss to limit the reward function, which makes the edited image retain most of the features of the original image. This also resulted in edited images that did not fulfill the instructions well, as evidenced by low CLIP-T scores.

\begin{table}[h]
  \caption{Ablation study. We attempt Attn-Only, Clip-Only, and our InstructRL4Pix to edit complex images on MagicBrush test set. Attn-Only speaks only of Attention loss as the reward function of the model. Clip-Only uses only Clip Loss as a reward function. And our InstructRL4Pix uses the combination of attention loss and Clip Loss as a reward function to accurately edit images while preserving original image features.}
  \label{tab:ablation}
  \small 
  \setlength{\tabcolsep}{4pt} 
  \begin{tabular}{@{}cccccl@{}}
    \toprule
    Methods &L1$\downarrow$ &L2$\downarrow$&CLIP-T$\uparrow$&SSIM$\uparrow$&PSNR$\uparrow$\\
    \midrule
    Attn-Only & 0.0601& 0.0184&0.2380&0.7545&20.0518\\
    Clip-Only & \textbf{0.0588}& \textbf{0.0167}&0.2362&0.7596&20.3118\\
    InstructRL4Pix & 0.0608& 0.0189&\textbf{0.2391}&\textbf{0.7978}&\textbf{24.2575}\\
    \bottomrule
  \end{tabular}
\end{table}
\subsection{Ablation Study}
\paragraph{Reward Function.}
PPO-Guided Image Editing has shown encouraging improvements in the precise editing of complex scenes. Now, we intend to verify the effectiveness of the reward function. Table 1 considers Attn-Only, Clip-Only, and our InstructRL4Pix.Attn-Only speaks only of Attention loss as the reward function of the model. In contrast, Clip-Only uses only Clip Loss as a reward function. The experimental results show that Clip-Only has the lowest L1 and L2, which is due to the fact that use Clip Loss as a reward function causes the model to more easily generate images that are similar to the original image and lose the editing function. Clip-T, SSIM, and PSNR metrics with attention loss only are not as good as those of our InstructRL4Pix, which indicates that the linear combination of attention loss and Clip Loss enables the model to successfully achieve fine editing of complex images while preserving the features of the original image.

\section{Conclusion}
In this work, we propose InstructRL4Pix, a method to train a diffusion model to generate images that are guided by the attention maps of the target object. Extensive experiments show that InstructRL4Pix is no longer restricted to the Prompt-to-Prompt framework and is able to accurately localize attention maps on complex images. We have used reinforcement learning algorithms firstly for instruction-based image editing, and we believe that an RL-guided image editing framework will contribute to future vision and language research.

\section{Acknowledgement}
This work was supported in part by the National Natural Science Foundation of China under Grant 62202174, in part by the Fundamental Research Funds for the Central Universities under Grant 2023ZYGXZR085, in part by the Basic and Applied Basic Research Foundation of Guangzhou under Grant 2023A04J1674, and in part by the Guangdong Provincial Key Laboratory of Human Digital Twin under Grant 2022B1212010004.
{\small
\bibliographystyle{ieee_fullname}
\bibliography{egbib}
}
\newpage
\appendix
\section{Implementation Details}
\subsection{Model Architecture}
\paragraph{Proximal Policy Optimization (PPO)}
Proximal Policy Optimization (PPO) is a policy optimization algorithm designed for reinforcement learning tasks, particularly suited for environments with high-dimensional action spaces. PPO belongs to the class of actor-critic algorithms, where an actor learns to predict the optimal policy, and a critic evaluates the value function associated with that policy. The surrogate objective function for PPO is given by:
\begin{equation}
L_{PPO}(\theta) = \mathbb{E}_{\pi_\theta} \left[ \min \left( r_t(\theta) A_t, \text{clip}(r_t(\theta), 1-\epsilon, 1+\epsilon) A_t \right) \right]
\end{equation}
where $\theta$ are the policy parameters, $r_t(\theta) = \frac{\pi_\theta(a_t|x_t)}{\pi_{\theta_{old}}(a_t|x_t)}$ is the probability ratio, $A_t$ is an estimate of the advantage function, and $\epsilon$ is a clipping parameter. 

The objective is to minimize the surrogate objective $L_{PPO}(\theta)$ using gradient descent:
\begin{equation}
\theta_{\text{new}} = \arg\min_{\theta} L(\theta)
\end{equation}
The advantage function $A_t$ can be estimated using a variety of techniques, such as generalized advantage estimation (GAE), which combines rewards with value estimates.

After computing the surrogate objective and estimating the advantage function, the policy parameters are updated to improve the policy while ensuring the update remains within a certain threshold defined by the clipping parameter $\epsilon$.

\paragraph{Modeling editing model as Multi-Step MDPs}
Following DDPO, We relate the diffusion modeling based editing model multi-step denoising process to the MDP in the following way:
\begin{equation}
s_t \triangleq (\mathcal{V}, \mathcal{X}, t, x_t)
\end{equation}
the state of each step is defined as a tuple containing the input image, the instruction, the denoising time step, and the denoising result of the current time step.
\begin{equation}
\pi(a_t|s_t) \triangleq p_{\theta}(x_{t-1}|x_t, \mathcal{V}, \mathcal{X})
\end{equation}
The strategy is the conditional distribution of the next denoising result given the current state.
\begin{equation}
A_t \triangleq x_{t-1}
\end{equation}
The action is the result of the denoising in the next step.
\begin{equation}
P (s_{t+1} | s_t, a_t) \triangleq (\delta_\mathcal{V}, \delta_\mathcal{X}, \delta_{t-1}, \delta_{x_{t-1}})
\end{equation}
During the denoising process, the state transfer is deterministic after sampling the next denoising result $x_{t-1}$ . Therefore, the state transfer probability is represented here by four Dirac delta distributions.
\begin{equation}
\rho_0(s_0) \triangleq (p(\mathcal{V}, \mathcal{X}), \delta_T , \mathcal{N} (0, \mathbb{I}))
\end{equation}
For the initial state distribution, the condition variable obeys its prior distribution, and the time step $T$ is determined, and finally xt then obeys standard Gaussian noise.
\begin{equation}
R(s_t, a_t) \triangleq \left\{
	\begin{aligned}
		& r(x_0, \mathcal{V}, \mathcal{X})  && t = 0\\
		& 0  && otherwise
	\end{aligned}
	\right.
\end{equation}
During the denoising process, only the final denoising result receives a score based on the reward model mentioned above, while the reward values during the denoising process are all defined as 0.

\paragraph{Learning of InstructRL4Pix}
Algo.\ref{alg:InstructRL4Pix} leverages a policy network trained with Proximal Policy Optimization (PPO) to iteratively generate attention maps that guide the editing process. These attention maps are compared with ground truth maps, facilitating the alignment of edited regions with target objects specified in input instructions. InstructRL4Pix employs a reward function incorporating cosine similarity between attention maps and a clip loss to regulate image modifications, optimizing them for both perceptual quality and fidelity to input images. By formulating the image editing task as a Markov Decision Process (MDP) and utilizing PPO for policy gradient estimation, InstructRL4Pix achieves efficient training of the editing model, leading to improved performance and stability in generating visually coherent edited images.
\begin{table}[htbp]
    \centering
    \caption{Training Diffusion for Image Editing by Reinforcement Learning Algorithm}
    \label{alg:InstructRL4Pix}
    \begin{tabular}{@{}l@{}}
        \toprule
        \textbf{Algorithm} InstructRL4Pix \\
        \midrule
        \textbf{Input:} Policy network parameters $\theta$, initial state $s_0$ \\
        \textbf{Output:} Updated policy network parameters $\theta$ \\
        \textbf{Initialization:} Initialize $\theta$ \\
        \midrule
        \textbf{Training Procedure:} \\
        \midrule
        \textbf{for} epoch = 1 to num\_epochs \textbf{do} \\
        \quad Sample trajectories using current policy \\
        \quad \textbf{for} trajectory in trajectories \textbf{do} \\
        \quad \quad \textbf{for} $t$ = 1 to $T$ \textbf{do} \\
        \quad \quad \quad Compute advantage function $A_t$ \\
        \quad \quad \quad Compute ratio $\rho_t = \frac{\pi_{\theta}(a_t|s_t)}{\pi_{\theta_{\text{old}}}(a_t|s_t)}$ \\
        \quad \quad \quad Compute clipped surrogate objective $L_{\text{PPO}}$ \\
        \quad \quad \quad Update policy parameters $\theta$ using gradient ascent \\
        \quad \textbf{end for} \\
        \quad Update old policy parameters $\theta_{\text{old}} = \theta$ \\
        \textbf{end for} \\
        \bottomrule
    \end{tabular}
\end{table}
\begin{table*}[htbp]
    \centering
    \caption{Training detials for IntructRL4Pix}
    \label{tab:Training detials for IntructRL4Pix}
\begin{tabular}{lll}
\toprule
\textbf{Parameter} & \textbf{Description} & \textbf{Value} \\
\midrule
config.seed & Seed for random number generation & 42 \\
config.logdir & Top-level logging directory for checkpoint saving & "logs" \\
config.num\_epochs & Number of epochs to train for & 200 \\
config.save\_freq & Number of epochs between saving model checkpoints & 50 \\
config.num\_checkpoint\_limit & Number of checkpoints to keep before overwriting old ones & 5 \\
config.mixed\_precision & Mixed precision training & "no" \\
config.allow\_tf32 & Allow tf32 on Ampere GPUs & True \\
config.resume\_from & Resume training from a checkpoint & "" \\
config.use\_lora & Whether or not to use LoRA & True \\
config.sample.num\_steps & Number of sampler inference steps & 50 \\
config.sample.eta & Eta parameter for the DDIM sampler & 1.0 \\
config.sample.guidance\_scale & Classifier-free guidance weight & 5.0 \\
config.sample.batch\_size & Batch size for sampling & 1 \\
config.sample.num\_batches\_per\_epoch & Number of batches to sample per epoch & 2 \\
config.train.batch\_size & Batch size for training & 1 \\
config.train.use\_8bit\_adam & Whether to use the 8bit Adam optimizer & True \\
config.train.learning\_rate & Learning rate & $2 \times 10^{-4}$ \\
config.train.adam\_beta1 & Adam beta1 & 0.9 \\
config.train.adam\_beta2 & Adam beta2 & 0.999 \\
config.train.adam\_weight\_decay & Adam weight decay & $1 \times 10^{-4}$ \\
config.train.adam\_epsilon & Adam epsilon & $1 \times 10^{-8}$ \\
config.train.gradient\_accumulation\_steps & Number of gradient accumulation steps & 1 \\
config.train.max\_grad\_norm & Maximum gradient norm for gradient clipping & 1.0 \\
config.train.num\_inner\_epochs & Number of inner epochs per outer epoch & 1 \\
config.train.cfg & Whether or not to use classifier-free guidance during training & True \\
config.train.adv\_clip\_max & Clip advantages to the range [-adv\_clip\_max, adv\_clip\_max] & 5 \\
config.train.clip\_range & The PPO clip range & $1 \times 10^{-4}$ \\
config.train.timestep\_fraction & The fraction of timesteps to train on & 1 \\
config.per\_prompt\_stat\_tracking.buffer\_size & Number of reward values to store in the buffer & 16 \\
config.per\_prompt\_stat\_tracking.min\_count & Minimum number of reward values & 16 \\
\bottomrule
\end{tabular}
\end{table*}
\subsection{Training Details}
Tab \ref{tab:Training detials for IntructRL4Pix} presents configuration parameters for InstructRL4Pix training process. Each row represents a parameter, including its name, description, and corresponding value or option. The description provides detailed explanations of each parameter, while the value column displays the actual numerical values or options adopted in the configuration. These parameters cover crucial settings such as random seed, number of epochs, model checkpoint saving frequency, learning rate, and more,

\begin{figure*}
  \includegraphics[width=\textwidth]{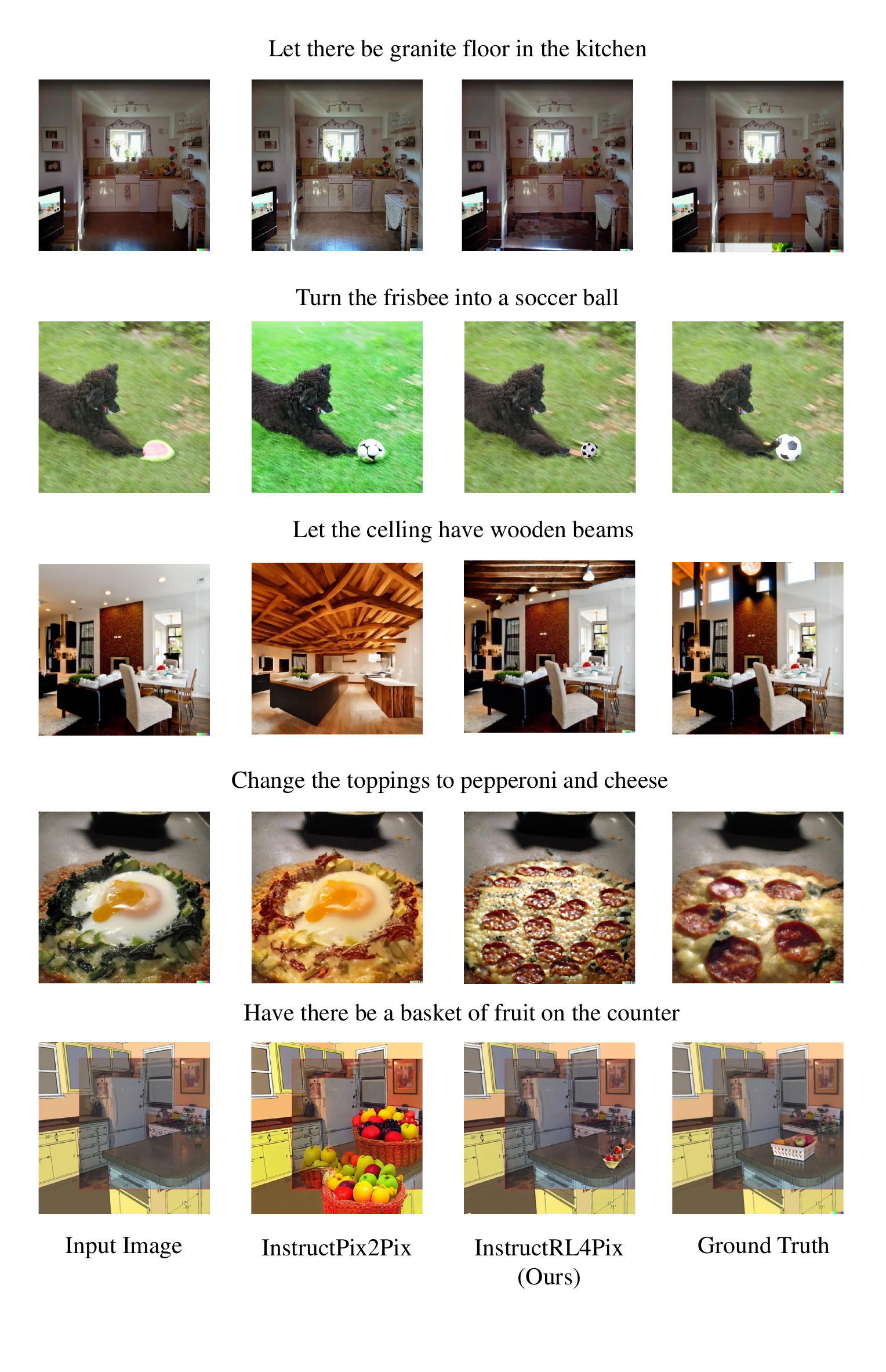}
  \caption{}
  \label{fig:examples}
\end{figure*}
\subsection{Evaluation Details}
We use 5 metrics to evaluate the model: L1, L2, CLIP-T,SSIM, PSNR. 
\paragraph{L1}
L1 measures the absolute pixel-wise differences between the predicted values and the ground truth.
It's calculated as the mean of the absolute differences between corresponding pixels in the predicted and ground truth images.
\[
L1 = \frac{1}{N} \sum_{i=1}^{N} |p_i - g_i|
\]
Here, $p_i$ and $g_i$ are the pixel values of the predicted and ground truth images respectively, and $N$ is the total number of pixels.
\paragraph{L2}
L2 measures the squared differences between the predicted values and the ground truth.
It's calculated as the mean of the squared differences between corresponding pixels in the predicted and ground truth images.
\[
L2 = \frac{1}{N} \sum_{i=1}^{N} (p_i - g_i)^2
\]

\paragraph{CLIP-T}
CLIP-T evaluates how well a model understands the semantics of the image by computing the similarity between image and text embeddings using the CLIP model.
Implementation might involve using CLIP embeddings and measuring cosine similarity between the embeddings of images and corresponding text descriptions.
\paragraph{SSIM}
SSIM (Structural Similarity Index) is a metric used to measure the similarity between two images, considering aspects like luminance, contrast, and structure. It evaluates how perceptually similar the images are, taking into account human visual perception. By comparing the brightness, contrast, and structural information at different scales, SSIM provides a comprehensive similarity measure ranging from 0 to 1, where a value closer to 1 indicates higher similarity. This metric is widely employed in image quality assessment and optimization tasks, aiding in tasks such as image compression, denoising, and enhancement. However, it's essential to acknowledge that SSIM has its limitations and might require adjustments based on specific application requirements.

\paragraph{PSNR}
PSNR measures the quality of a reconstructed signal.
It compares the maximum possible power of a signal with the power of corrupting noise.
\[
PSNR = 10 \cdot \log_{10}\left(\frac{{\text{{max pixel value}}^2}}{{\text{{MSE}}}}\right)
\]
Here, MSE is the Mean Squared Error calculated earlier.

\subsection{More Results}
Figure \ref{fig:examples} shows more examples for results of our model.

\end{document}